\documentclass[10pt,twocolumn,letterpaper]{article}

\usepackage[pagenumbers]{cvpr} 
\usepackage{extarrows}
\usepackage{color} 
\usepackage{xcolor}
\usepackage{makecell}
\usepackage{amsmath,amsfonts}
\usepackage{algorithmic}
\usepackage{array}
\usepackage{graphicx}
\usepackage{stfloats}
\usepackage{multicol}
\usepackage{multirow}
\usepackage{hhline}
\usepackage{colortbl}
\usepackage{bm}
\usepackage{url}
\usepackage{enumitem}
\usepackage{arydshln}
\usepackage{bbding}
\usepackage{utfsym}
\usepackage{makecell}
\usepackage{booktabs} 

\definecolor{cvprblue}{rgb}{0.21,0.49,0.74}
\usepackage[pagebackref,breaklinks,colorlinks,allcolors=cvprblue]{hyperref}
\usepackage{enumitem}
\def\paperID{*****} 
\def\confName{CVPR}
\def\confYear{2025}

\title{
CIR-CoT: 
Towards Interpretable Composed Image Retrieval via\\ End-to-End Chain-of-Thought Reasoning}

\author{Weihuang Lin\footnotemark[1],\:
Yiwei Ma\footnotemark[1],\:
Jiayi Ji,\:
Xiaoshuai Sun\footnotemark[2],\:
Rongrong Ji\\
Key Laboratory of Multimedia Trusted Perception and Efficient Computing, \\ Ministry of Education of China, Xiamen University, 361005, P.R. China \\
}
\begin{document}
\maketitle

\renewcommand{\thefootnote}{\fnsymbol{footnote}}
\footnotetext[1]{These authors contributed equally to this work.} 
\footnotetext[2]{The corresponding author.}

\begin{abstract}
Composed Image Retrieval (CIR), which aims to find a target image from a reference image and a modification text, presents the core challenge of performing unified reasoning across visual and semantic modalities. While current approaches based on Vision-Language Models (VLMs, e.g., CLIP) and more recent Multimodal Large Language Models (MLLMs, e.g., Qwen-VL) have shown progress, they predominantly function as ``black boxes." This inherent opacity not only prevents users from understanding the retrieval rationale but also restricts the models' ability to follow complex, fine-grained instructions. To overcome these limitations, we introduce CIR-CoT, the first end-to-end retrieval-oriented MLLM designed to integrate explicit Chain-of-Thought (CoT) reasoning. By compelling the model to first generate an interpretable reasoning chain, CIR-CoT enhances its ability to capture crucial cross-modal interactions, leading to more accurate retrieval while making its decision process transparent. Since existing datasets like FashionIQ and CIRR lack the necessary reasoning data, a key contribution of our work is the creation of structured CoT annotations using a three-stage process involving a caption, reasoning, and conclusion. Our model is then fine-tuned to produce this structured output before encoding its final retrieval intent into a dedicated embedding. Comprehensive experiments show that CIR-CoT achieves highly competitive performance on in-domain datasets (FashionIQ, CIRR) and demonstrates remarkable generalization on the out-of-domain CIRCO dataset, establishing a new path toward more effective and trustworthy retrieval systems.
\end{abstract}    
\section{Introduction}
\label{sec:intro}

\begin{figure}

  \centering
  \includegraphics[width=1.\columnwidth]{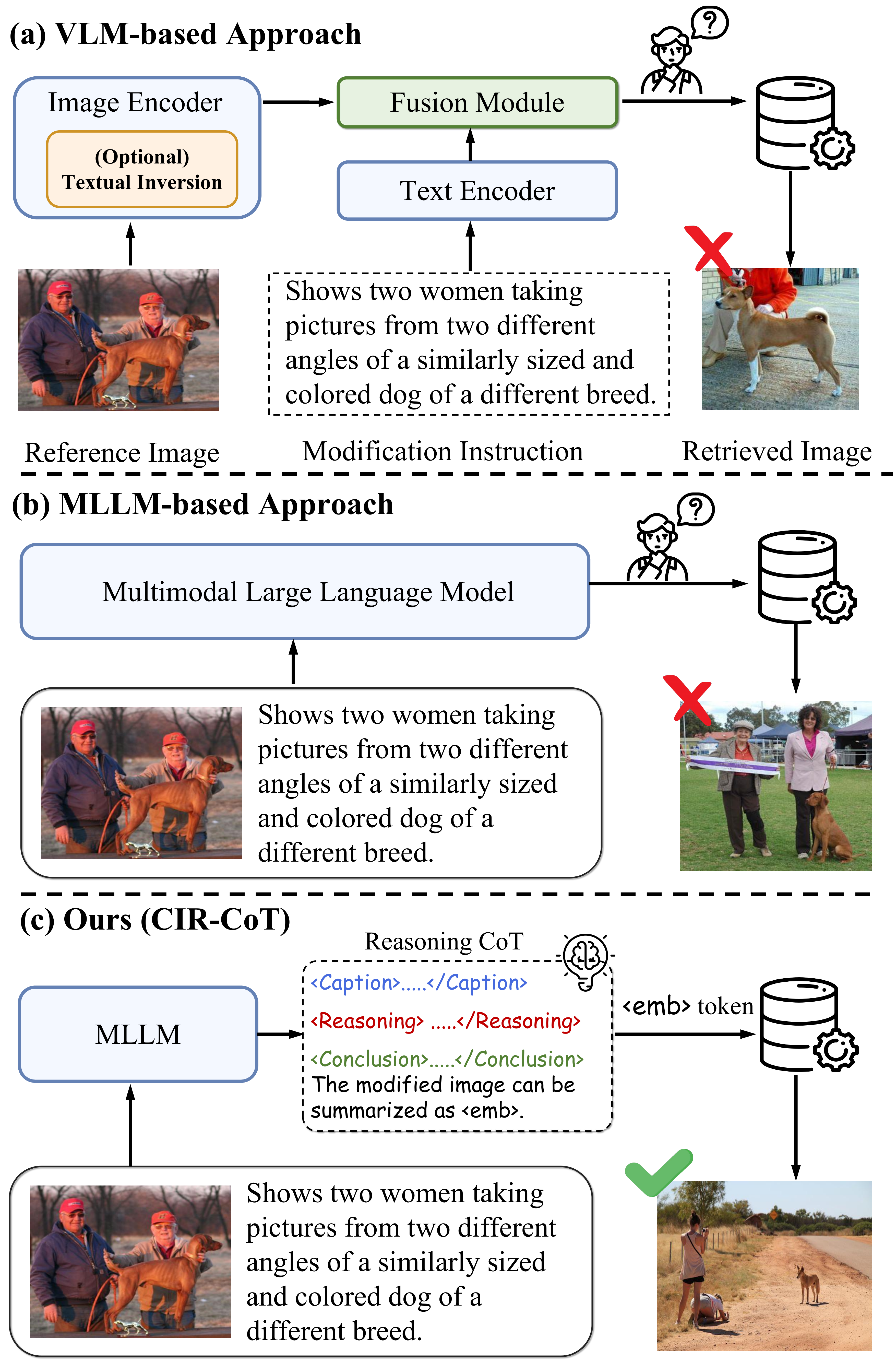}
  \vspace{-2em}
  \caption{Comparison of three retrieval approaches: 
(a) VLM-based method; 
(b) MLLM-based method (treating the MLLM as an encoder); 
(c) our CIR-CoT approach, enhanced with Chain-of-Thought reasoning for more accurate image retrieval.}
  \label{fig:intro}
\vspace{-1em}
  
\end{figure}

Composed Image Retrieval (CIR) builds on traditional image retrieval~\cite{gordo2016deep,liu2016deepfashion,liu2018attentive,yang2022video} by allowing users to provide a reference image along with a modification instruction. This flexibility makes CIR particularly useful for applications like e-commerce product search, where users often look for visually similar items with specific variations.
To retrieve the desired target image, the key challenge in CIR task lies in reasoning over the visual content of the reference image and the semantics of the modification instruction in a unified manner. As a challenging multimodal retrieval task, CIR has attracted increasing attention in both academia and industry.

To address the CIR task, current research primarily follows two mainstream approaches. 
The first category builds on the success of Vision-Language Models (VLMs)~\cite{radford2021learning,jia2021scaling}, as shown in Fig.~\ref{fig:intro} (a). Specifically, some methods~\cite{anwaar2021compositional,chen2020image,liu2021image,levy2024data,liu2023candidate} encode the reference image and the modification text separately using VLM encoders, and perform feature fusion to retrieve the target image.
Other methods~\cite{saito2023pic2word,bai2023sentence} go beyond such straightforward fusion by first transforming the reference image into a textual embedding using mechanisms like textual inversion~\cite{gal2022image}, which is then combined with the modification instruction. While this strategy enhances the model’s ability to interpret complex user intent, the gains remain limited. This highlights the need for stronger semantic reasoning across modalities.
Recently, Multimodal Large Language Models (MLLMs), such as LLaVA~\cite{liu2023visual,liu2024llavanext} and Qwen-VL~\cite{wang2024qwen2,bai2025qwen2}, have gained popularity for their strong multimodal reasoning capabilities. Inspired by this progress, various studies~\cite{jiang2024e5,liu2025lamra,zhang2024gme} explore the use of MLLMs for universal retrieval tasks.
Finetuning MLLMs specifically for the CIR task has only recently been attempted, as shown in Fig.~\ref{fig:intro} (b). Specifically, CIR-LVLM~\cite{sun2025leveraging} pioneers this direction, achieving strong performance in understanding user intent and aggregating hybrid-modality query features, thereby demonstrating the effectiveness and promise of MLLMs for CIR. 

Despite these advances, existing retrieval methods, including both VLM-based approaches and recent MLLM-based solutions, largely treat the model as a black box. In other words, users have little visibility into how the model reasons over hybrid-modality queries, which makes it difficult to verify the retrieved results.
An interpretable reasoning process is therefore essential, since it not only enables users to understand the rationale behind retrieval decisions but also guides the model to perform structured reasoning over multimodal inputs. Such reasoning allows the model to capture critical cross-modal interactions that might otherwise be overlooked, ultimately improving retrieval performance. As illustrated in Fig.~\ref{fig:intro} (c), our approach successfully retrieves the correct target image under a complex instruction, whereas prior methods fail and provide no interpretable rationale.

Therefore, we propose CIR-CoT, an end-to-end retrieval-oriented MLLM that performs explicit reasoning over interleaved multimodal inputs. The main challenge in training CIR-CoT is the lack of structured reasoning annotations in existing CIR datasets, such as FashionIQ~\cite{wu2021fashion} and CIRR~\cite{liu2021image}, which only provide basic image–instruction pairs. Inspired by LLaVA-CoT~\cite{xu2024llava}, we extend existing datasets with enriched annotations. Instead of generating a direct reasoning chain, we employ a multistage reasoning approach to structure the annotations. Specifically, we leverage the powerful open-source multimodal model Qwen2.5-VL-72B~\cite{bai2025qwen2} to produce three-stage annotations:
\begin{enumerate}
    \item \textbf{Caption}: Extracting detailed visual features from the reference image.
    \item \textbf{Reasoning}: Deliberating on how to integrate the reference image and the modification instruction.
    \item \textbf{Conclusion}: Deriving a description of the target image that should be retrieved, based on the reasoning process.
\end{enumerate}
To ensure the accuracy of the annotations, we extract the Conclusion from each sample and conduct a multi-expert review, comparing it against the correct target image in the dataset and filtering out any samples with inconsistent or incorrect annotations.

Based on the annotated dataset, we train CIR-CoT in two stages. In the first stage, the model is pretrained on the pure-text NLI dataset~\cite{gao2021simcse} to enhance its summarization ability, enabling it to effectively compress information into the newly introduced \texttt{<emb>} token. In the second stage, we finetune the model on the extended CIRR and FashionIQ datasets. The goal is to guide the model to first produce a structured Chain-of-Thought reasoning output, and then encode the retrieval intent into the \texttt{<emb>} token embedding, which acts as the semantic representation for retrieval. By enforcing a structured reasoning process, the model is encouraged to explicitly examine cross-modal interactions, which improves its ability to capture fine-grained details and better interpret complex user intent. Meanwhile, this process also makes the retrieval procedure more transparent to users, providing interpretable rationales and moving beyond the traditional black-box paradigm of retrieval.

To evaluate the effectiveness of CIR-CoT, we conduct experiments on in-domain datasets, FashionIQ and CIRR, as well as the out-of-domain dataset CIRCO~\cite{baldrati2023zero}. The results demonstrate that CIR-CoT not only achieves strong performance on in-domain benchmarks but also exhibits remarkable generalization ability on out-of-domain data.
\begin{figure*}

  \centering
  \includegraphics[width=2\columnwidth]{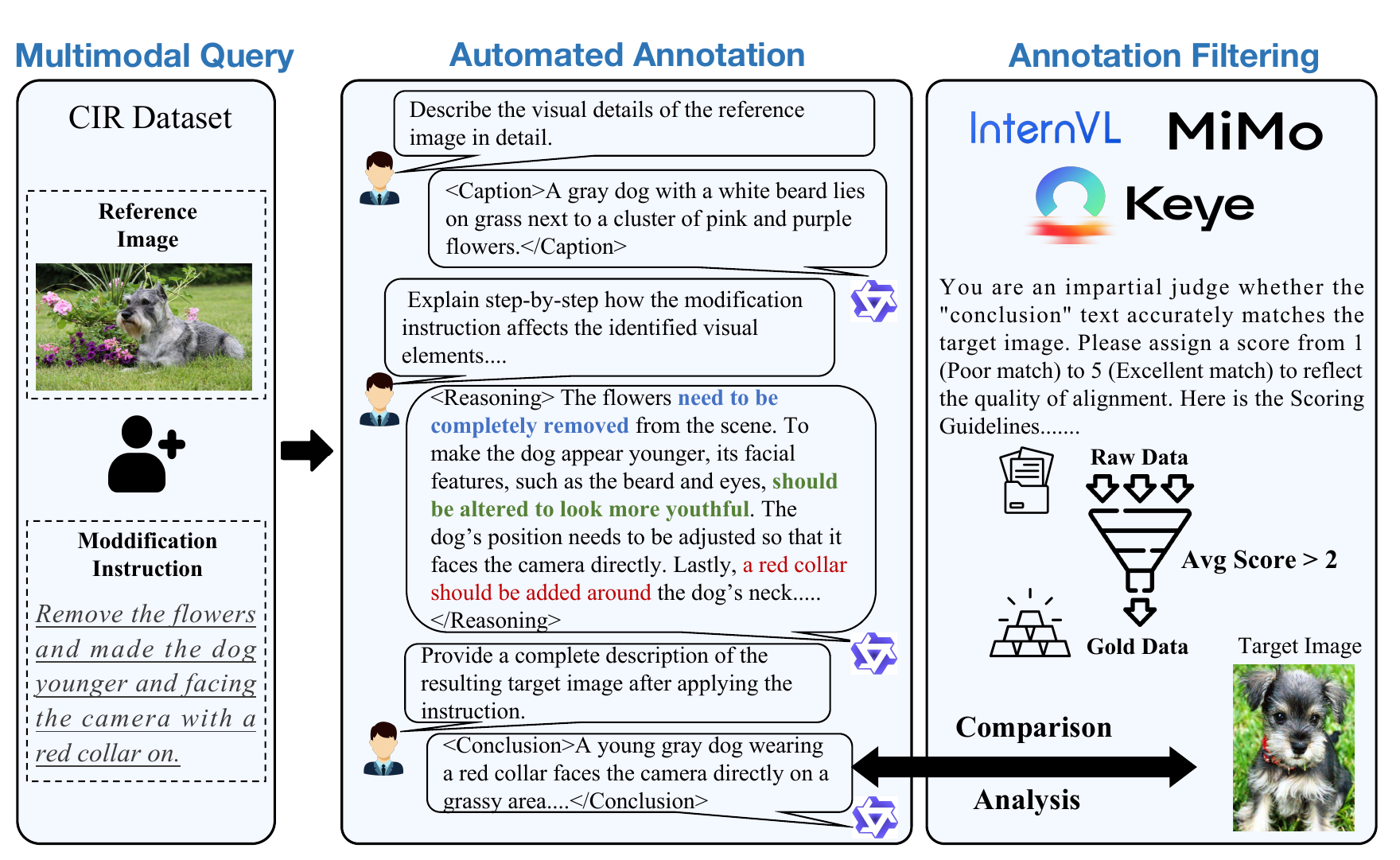}
  \caption{The pipeline for constructing CoT training data. A multimodal query is processed through automated annotation to produce reasoning-augmented descriptions, followed by MLLM-based evaluation for quality control.}
  \label{fig:data}
\vspace{-1.2em}
  
\end{figure*}

In summary, the contributions of this paper are threefold:
\begin{itemize}[leftmargin=2em]
    \item We construct \textbf{structured CoT-annotated datasets} by extending FashionIQ and CIRR with structured CoT annotations, providing valuable resources for reasoning-oriented CIR research.
    
    \item We propose \textbf{CIR-CoT}, the first end-to-end retrieval-oriented MLLM that incorporates explicit Chain-of-Thought reasoning, enabling interpretable and more accurate compositional image retrieval.
    
    \item We conduct comprehensive experiments on both in-domain datasets (FashionIQ, CIRR) and the out-of-domain dataset (CIRCO), demonstrating that CIR-CoT achieves competitive retrieval performance and strong generalization ability.

\end{itemize}

\section{Related Work}
\label{sec:related}
\subsection{Composed Image Retrieval}

Recent advances in Vision–Language Models (VLMs)~\cite{jia2021scaling,Li2023BLIP2BL} have laid a strong foundation for compositional image retrieval. Building on these models, most contemporary CIR approaches develop various adaptation strategies to tailor them to the retrieval task. Specifically, some methods~\cite{Levy2023DataRA,Liu2023CandidateSR,Anwaar2020CompositionalLO,Chen_2020_CVPR} adopt an early-fusion strategy, where the text and image features are first extracted separately using unimodal encoders and then fused to form a joint query representation, which is subsequently matched against candidate features. The main limitation of such early-fusion approaches lies in their inability to accurately align fine-grained visual details with user intent during feature fusion. To address this issue, another line of work~\cite{bai2023sentence,gal2022image,saito2023pic2word,Tang2023ContextI2WMI} transforms the reference image into a word embedding via textual inversion, concatenates it with the query text to form an enhanced textual feature, and then performs text-to-image retrieval. Despite their effectiveness, the reliance on text encoders limits these methods’ ability to faithfully interpret and retrieve images according to complex user intent. Consequently, a recent work, CIR-LVLM~\cite{sun2025leveraging}, attempts to finetune MLLMs to better capture user intent by directly encoding multimodal inputs and retrieving the target image accordingly. Leveraging the strong comprehension ability of MLLMs, this approach achieves promising results. Unlike prior work, CIR-CoT fully exploits MLLMs by (i) generating explicit, human-readable reasoning that makes retrieval transparent rather than black-box, and (ii) encoding the reasoned user intent as a retrieval representation, yielding stronger performance.

\subsection{Multimodal Large Language Models}
Large Language Models (LLMs)~\cite{chiang2023vicuna,touvron2023llama,zheng2023judging,team2023internlm,openai2023gpt,meta2024introducing,bi2024deepseek,yang2024qwen2} have recently achieved remarkable progress, attracting broad research interest due to their strong reasoning and generation abilities. Building on this success, researchers have extended LLMs to handle visual inputs, which has driven rapid advances in Multimodal Large Language Models (MLLMs)~\cite{liu2023visual,bai2023qwen,zhu2023minigpt,lu2024deepseek,liu2024oryx,li2024mini}. Recent studies have shown that MLLMs excel in diverse vision tasks. Notably, some approaches~\cite{Lai2023LISARS, Lin2025HRSegHV} employ MLLMs for segmentation, marking a departure from the conventional VQA paradigm. However, MLLMs tend to exhibit hallucinations when performing complex tasks and often underutilize visual information.
To address these challenges, some approaches~\cite{Qiao2024PrismAF,Cesista2024MultimodalSG,Chu2023NavigateTE} leverage Chain-of-Thought (CoT) prompting, which decomposes a question into a series of reasoning steps and constructs a chain to guide the model in generating solutions to complex problems. This process significantly enhances the reasoning capabilities of MLLMs. Although direct CoT approaches are effective, later methods~\cite{Xu2024LLaVACoTLV} demonstrated that the proposed structured CoT significantly outperforms direct CoT, further enhancing the reasoning capabilities of MLLMs. 
Building on the developments mentioned above, CIR-CoT is the first approach to apply the structured CoT reasoning capabilities of MLLMs to the CIR task. Its goal is to stimulate fine-grained reasoning in MLLMs over different user inputs and to infer user intent, thereby improving retrieval performance.

\begin{figure*}

  \centering
  \includegraphics[width=2\columnwidth]{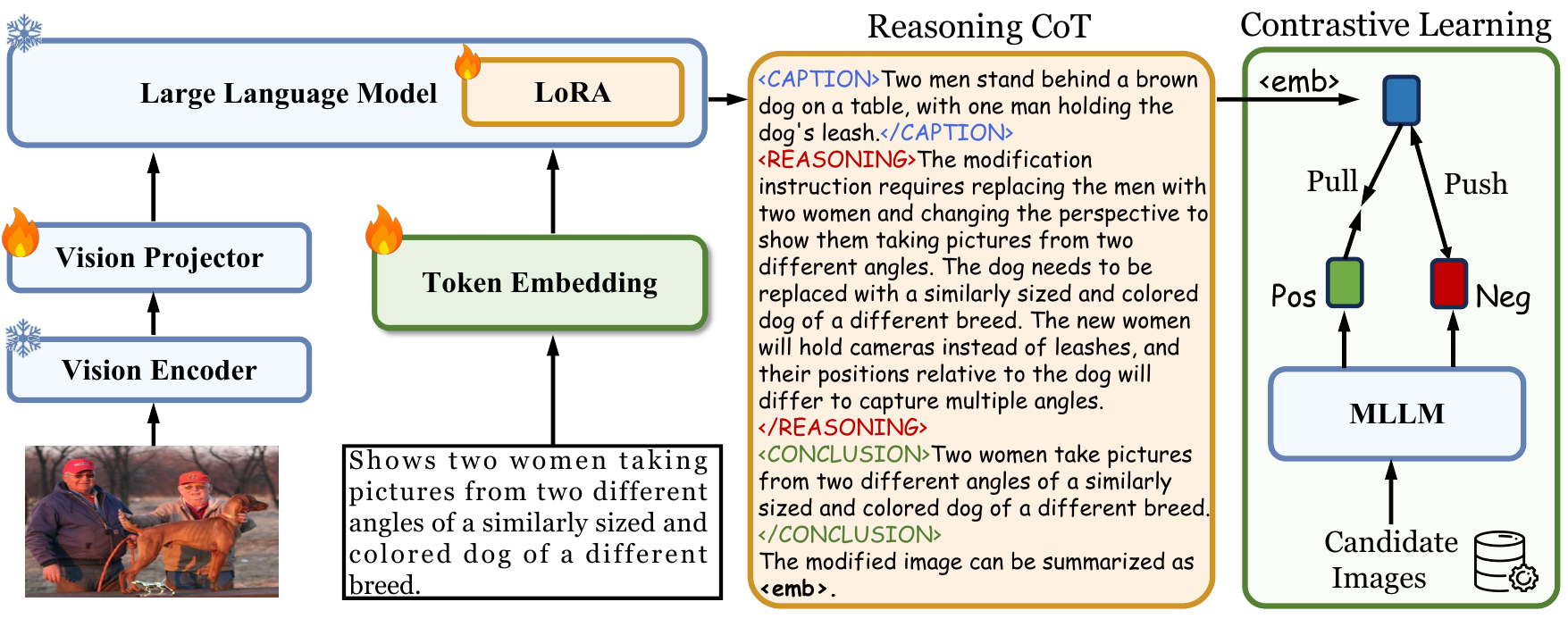}
  \vspace{-1em}
  \caption{Overview of the proposed baseline CIR-CoT. The method leverages MLLMs to generate reasoning chains for the target image and obtain its embedding token \texttt{<emb>}, followed by contrastive learning to improve retrieval.}
  \label{fig:method}
  \vspace{-1.2em}
  
\end{figure*}
\section{Method}
In this section, We first present the problem formulation of the Composed Image Retrieval task. This is followed by a description of the procedure for constructing a CoT-annotated dataset in Sec.\ref{method_one}, which provides the foundation for reasoning-aware retrieval. Subsequently, we present the architecture of our proposed CIR-CoT model in Sec.\ref{method_two}, highlighting how structured chain-of-thought reasoning is integrated into the retrieval framework. Finally, the training strategy and objectives are introduced in Sec.~\ref{method_three}.

\noindent \textbf{Problem Formulation.}
Let $\mathcal{D} = \{(I_i, M_i, T_i)\}_{i=1}^{N}$ denote the CIR dataset, where $r_i$ is the reference image, $M_i$ is the modification instruction, and $T_i$ is the corresponding target image. Given a reference image $I_i$ and a modification instruction $M_i$, the goal of Composed Image Retrieval (CIR) is to learn a retrieval function
\begin{equation}
f(I_i, M_i) \rightarrow \hat{T}_i \in \mathcal{D}_c,
\end{equation}
where $\mathcal{D}_c$ denotes the set of candidate images in the database, and $\hat{T}_i$ is the image retrieved by the model in response to the query $(I_i, M_i)$. The learning process aims to maximize the accuracy of the matching such that $\hat{T}_i = T_i$.

This formulation emphasizes the challenge of capturing the compositional relationship between the reference image and the modification instruction, requiring the model to reason over both visual and textual modalities to retrieve the correct target.


\subsection{Data generation}
\label{method_one}
Fig.~\ref{fig:data} presents the overall procedure for structured CoT annotation. We begin by extracting the multimodal query, the reference image, and the modification instruction from the FashionIQ and CIRR datasets, which provide diverse and realistic benchmarks for compositional retrieval. These elements are then automatically annotated to generate structured reasoning traces that decompose the query into interpretable steps. To ensure the reliability and quality of the annotated data, we further employ multiple MLLMs as expert judges to evaluate the generated reasoning and remove any instances that are inconsistent or logically unsound.

More specifically, we employ Qwen2.5-VL 72B to generate the automated annotation in a single inference pass, which is divided into three stages:
\begin{enumerate}
\item \textbf{Caption Stage:} The model is guided to focus on the visual details of the reference image, capturing all visible objects, attributes, and contextual elements. This stage ensures that fine-grained information is preserved and prevents the model from overlooking important visual details.

\item \textbf{Reasoning Stage: }This is the core stage, where the model is instructed to provide a chain-of-thought explanation. Concretely, the model executes the following steps:
\begin{itemize}[leftmargin=2em]
  \item \textbf{Comprehend the instruction:} extract the core visual goal, i.e., what to add, remove, or change.
  \item \textbf{Align with the reference image:} map the instruction's intent to existing objects, attributes, and spatial relationships in the image.
  \item \textbf{Determine concrete visual adjustments:} decide whether the change requires addition, removal, repositioning, attribute modification, and identify the specific target entities.
  \item \textbf{Form a clear reasoning chain:} present a step-by-step logical explanation of how the adjustments transform the reference image into the target image, and explain why each modification is necessary.
\end{itemize}
This process ensures that the reasoning explicitly ties the user's modification intent to fine-grained visual details and yields interpretable, stepwise transformation traces.
\item \textbf{Conclusion Stage:} Based on the preceding reasoning, the model produces a clear and comprehensive description of the resulting target image after applying the instruction. This final description serves as the semantic representation of the image to be retrieved.

\end{enumerate}

In addition, after the automated annotation, we adopt \textbf{the Annotation Filtering}, following the practice in~\cite{chen2024mllm}, to ensure annotation quality and mitigate hallucinations. Specifically, the content generated in the Conclusion Stage is extracted and compared against the ground-truth target image. Multiple MLLMs, including recent advanced models such as InternVL3~\cite{zhu2025internvl3}, MiMo-VL~\cite{Yue2025MiMoVLTR}, and Keye-VL~\cite{team2025kwai}, are employed to assign multi-level scores that assess consistency. Finally, annotations with significant discrepancies are discarded.
\subsection{CIR-CoT Architecture}
\label{method_two}

In Fig.~\ref{fig:method}, we present an overview of the proposed CIR-CoT framework. 
The architecture consists of a vision encoder $f_{\text{VE}}$, a projection module $f_{\text{proj}}$, and a large language model $f_{\text{LLM}}$. 
Given a reference image $I$ and a modification instruction $M$, the vision encoder first extracts visual features:
\begin{equation}
v = f_{\text{VE}}(I),
\end{equation}
where $v$ denotes the visual representation of the reference image. 
These features are then mapped into the language embedding space by the projection layer:

\begin{equation}
\tilde{v} = f_{\text{proj}}(v),
\end{equation}
which produces $\tilde{v}$ as the language-aligned visual embedding to be consumed by the LLM.

The instruction $M$ is tokenized and embedded into $\tilde{m}$, and concatenated with the projected visual feature $\tilde{v}$. 
The fused sequence is then fed into the LLM backbone, which autoregressively generates a sequence of output tokens:
\begin{equation}
\hat{y}_{\text{txt}} = f_{\text{LLM}}([\tilde{v}, \tilde{m}]) = (y_1,\dots,y_T),
\end{equation}
where $T$ denotes the sequence length and each $y_t$ corresponds to a generated token. 
The generation process follows the standard conditional factorization:
\begin{equation}
p_\theta(\hat{y}_{\text{txt}}\mid \tilde{v}, \tilde{m}) = \prod_{t=1}^{T} p_\theta(y_t \mid y_{<t}, \tilde{v}, \tilde{m}),
\end{equation}

By design, $\hat{y}_{\text{txt}}$ contains a structured chain-of-thought (CoT) reasoning trace that explicitly decomposes the query into interpretable steps:
\begin{equation}
\mathcal{R}(I, M) = \{s_1, s_2, \dots, s_K\},
\end{equation}
where each $s_k$ denotes a reasoning step.

Beyond generating the reasoning sequence, CIR-CoT appends a special token \texttt{<emb>} at the end of the output to summarize the target image representation. 
We extract the last-layer hidden state corresponding to this token as the target image embedding:
\begin{equation}
e_q = f_{\text{LLM}}^{\text{last}}(\texttt{<emb>}).
\end{equation}
This embedding $e_q$ serves as a compact representation of the user’s intent and captures the semantic characteristics of the target image.

\subsection{Training Strategy and Objectives}
\label{method_three}
We adopt a two-stage training strategy to adapt the MLLM backbone for compositional image retrieval. 
The motivation is that general-purpose MLLMs are primarily optimized for text generation rather than retrieval, and thus cannot directly produce compact embeddings suitable for matching tasks. 
To address this, we progressively guide the model to learn how to compress user input semantics into the designated \texttt{<emb>} token.

\textbf{Stage 1: Textual Embedding Pretraining.} 
Inspired by~\cite{jiang2024e5}, we first pretrain the model on a large-scale textual dataset, specifically the natural language inference (NLI) dataset.
During training, we design a simple prompt: 
\textit{``Summarize the above sentence in one word: \texttt{<emb>}''}, 
which encourages the LLM to encode the essential semantics of the input into the \texttt{<emb>} token. 
This stage equips the model with the ability to perform semantic compression in the purely textual domain.

\textbf{Stage 2: Multimodal CoT Adaptation.} 
After pretraining, we further fine-tune the model on the CoT-annotated multimodal dataset constructed in Sec.~\ref{method_one}. 
This stage transfers the semantic compression ability to multimodal queries by training the model to not only generate structured reasoning traces but also summarize the final target image into the \texttt{<emb>} token, which serves as the target image embedding for retrieval.

To optimize the retrieval ability, the model is trained end-to-end with a combination of the text generation loss and the InfoNCE loss.

During training, the autoregressive generation of reasoning traces is supervised using a cross-entropy loss:
\begin{equation}
\mathcal{L}_{\text{txt}} = \mathbf{CE}(y_{\text{txt}}, \hat{y}_{\text{txt}}),
\end{equation}
where $y_{\text{txt}}$ denotes the ground-truth sequence, and $\hat{y}_{\text{txt}}$ is the predicted sequence generated by the model. This objective ensures that the LLM produces faithful and interpretable reasoning sequences aligned with the annotated CoT data.

To learn discriminative embeddings for retrieval, we adopt the InfoNCE loss~\cite{oord2018representation}. Given a batch of $N$ query–image pairs $\{(e_q^j, e_i^j)\}_{j=1}^N$, we aim to align each query with its corresponding target image while pushing it away from negatives within the batch. 
The InfoNCE loss is defined as:
\begin{equation}
\mathcal{L}_{\text{InfoNCE}} = - \frac{1}{N} \sum_{j=1}^{N} 
\log \frac{\exp \left( \text{sim}(e_q^j, e_i^j) / \tau \right)}
{\sum_{k=1}^{N} \exp \left( \text{sim}(e_q^j, e_i^k) / \tau \right)},
\end{equation}
where $\text{sim}(\cdot,\cdot)$ denotes the cosine similarity function, and $\tau$ is a temperature hyperparameter.

The overall training objective combines the two losses:
\begin{equation}
\mathcal{L} = \lambda_{txt}\mathcal{L}_{txt} + \lambda_{\text{Info}}\mathcal{L}_{\text{InfoNCE}},
\end{equation}
where $\lambda_{txt}$ and $\lambda_{\text{Info}}$ are weighting coefficients that balance the two objectives.
\section{Experiments}

\subsection{Dataset and Evaluation Metric}

We evaluate CIR-CoT on three widely used CIR benchmarks: Fashion-IQ~\cite{wu2021fashion}, CIRR~\cite{liu2021image}, and CIRCO~\cite{baldrati2023zero}.  
Fashion-IQ focuses on the fashion domain with triplets drawn from web-crawled product images.  
CIRR provides a more general real-world setting and further includes a fine-grained subset with visually similar candidates, making retrieval particularly challenging.  
CIRCO is constructed from COCO images, offering large-scale distractors and multiple annotated ground-truth matches to alleviate the false negative issue in CIRR.  
For performance evaluation, we adopt Recall@K as the evaluation metric. Specifically, for CIRR, we report Recall@1, 5, 10, and 50 to measure global retrieval accuracy, as well as Recall$_\text{subset}$@1, 2, and 3 to capture fine-grained discrimination within visually similar candidates. For Fashion-IQ, we follow the standard protocol and provide Recall@10 and Recall@50 results across the three fashion categories. For CIRCO, we use mean average precision at rank $k$ (mAP@k) to account for multiple valid ground-truth targets in the retrieval set.

\subsection{Implementation Details}

CIR-CoT is built upon Qwen2.5-VL-7B as the backbone, with LoRA applied for efficient parameter-efficient fine-tuning. All experiments are conducted on 8 NVIDIA A800 GPUs. In the first stage of pretraining, the model is optimized on the NLI dataset using only the $\mathcal{L}_{\text{InfoNCE}}$ objective, in order to encourage the backbone to develop retrieval-oriented representations. This stage is trained for 2 epochs with a batch size of 768 and a learning rate of $3\times10^{-4}$. In the second stage of finetuning, the model is trained on our CoT-annotated extensions of Fashion-IQ and CIRR (Sec.~\ref{method_one}). For CIRR, we train the model for up to 3 epochs with a global batch size of 320 and a learning rate of $2\times10^{-4}$. For Fashion-IQ, we adopt the same maximum number of epochs 3 with a global batch size of 288 and a learning rate of $3\times10^{-4}$. Both $\lambda_{txt}$ and $\lambda_{\text{Info}}$ are set to 1.0.
\begin{figure*}
\vspace{-0.5em}

  \centering
  \includegraphics[width=2\columnwidth]{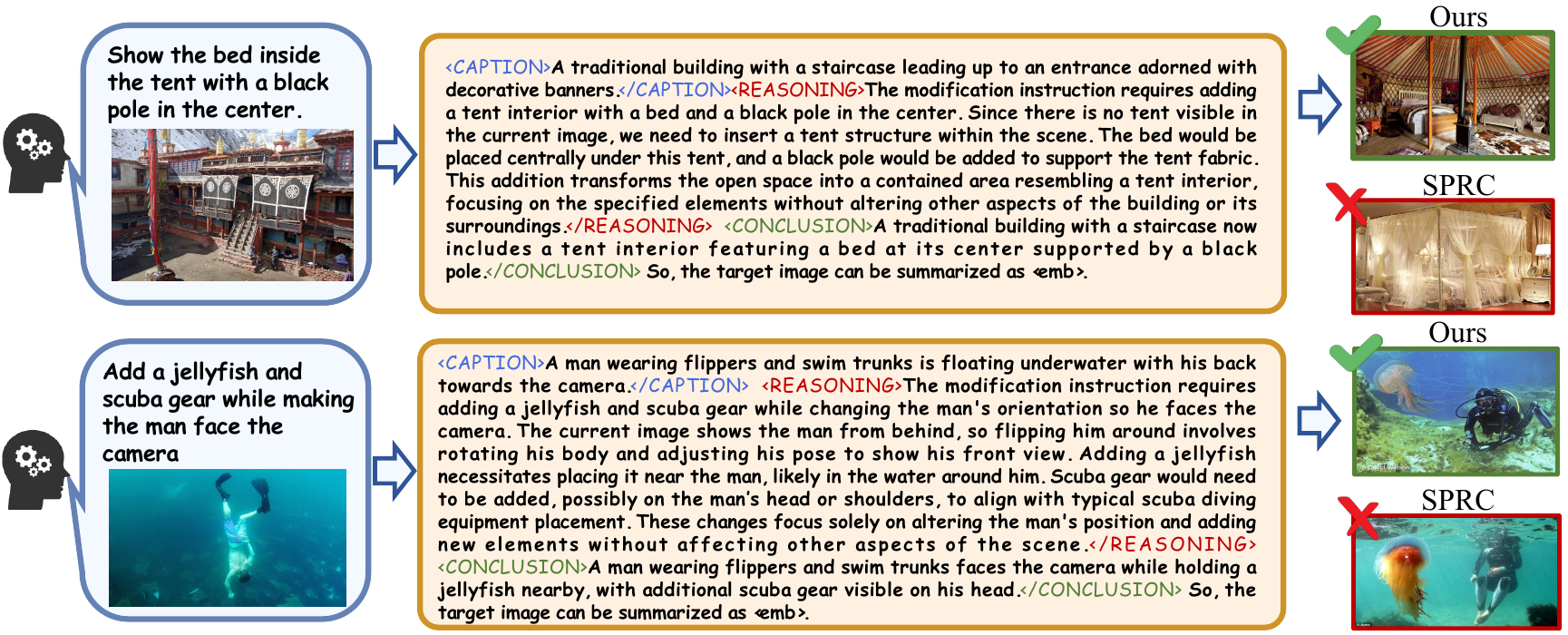}
  \vspace{-1em}
  \caption{Qualitative Results on CIRR dataset.}
  \label{fig:visual}
  
\end{figure*}
\subsection{Results on CIRR}

Table~\ref{tab:cirr} reports the performance comparison on the CIRR test set. 
CIR-CoT clearly outperforms all competing methods across most evaluation metrics. 
For instance, CIR-CoT (Full) achieves an R@1 of 55.06, surpassing recent strong baselines such as CCIN~\cite{tian2025ccin} (53.41) and QuRe~\cite{kwak2025qure} (52.22). 
On R@5, our method improves upon the previous best CCIN (84.05) by +1.42 points, reaching 85.47. 
A similar trend is observed on R@10, where CIR-CoT attains 92.60 compared to 91.17 from CCIN. 
Moreover, CIR-CoT achieves the highest overall average score of 82.49, which is +0.48 higher than the strongest prior method TME~\cite{li2025learning} (82.01).  
On the fine-grained subset evaluation, CIR-CoT delivers competitive results, obtaining the best R$_{subset}$@2 (92.92) and R$_{subset}$@3 (97.37).  We also report a variant, \textbf{\emph{CIR-CoT (Fast)}}, which considers efficiency. Since generating long reasoning chains inevitably slows down retrieval, we train a lighter version by retaining only the conclusion part of the CoT annotations. This reduces the number of tokens generated at inference time, resulting in faster retrieval while maintaining competitive performance (e.g., 54.31 R@1). More detailed efficiency analyses are provided in the supplementary material.

\begin{table*}[]
  \centering
  \setlength{\tabcolsep}{12pt}
  \vspace{-10pt}
      \caption{Performance comparison on the CIRR test set. The “Avg.” metric is computed as $(\text{R@5} + \text{R}_{subset}@1)/2$. }
        \vspace{-10pt}
    \small
    \begin{tabular}{lcccc ccc c}
    \toprule
    \multicolumn{1}{c}{\multirow{2}{*}{Method}} & \multicolumn{4}{c}{R@$k$}      & \multicolumn{3}{c}{R$_{subset}$@$k$} & \multirow{2}{*}{Avg.} \\
\cmidrule(lr){2-5}\cmidrule(lr){6-8} & k=1   & k=5   & k=10  & k=50  & k=1   & k=2   & k=3   &  \\
\midrule
    TG-CIR~\cite{wen2023target}~\textcolor{gray}{(ACM MM'23)} & 45.25  & 78.29  & 87.16  & 97.30  & 72.84  & 89.25  & 95.13  & 75.57  \\
    LIMN~\cite{wen2023self}~\textcolor{gray}{(TPAMI'24)} &43.64  & 75.37  & 85.42  & 97.04  & 69.01  & 86.22  & 94.19  & 72.19 \\
    SADN~\cite{wang2024semantic}~\textcolor{gray}{(ACM MM'24)} &44.27& 78.10& 87.71 &97.89& 72.71 &89.33& 95.38& 75.41\\
    DQU-CIR~\cite{wen2024simple}~\textcolor{gray}{(SIGIR'24)}& 46.22 &	78.17 &	87.64 &	97.81 	&70.92 &	87.69 &	94.68 &	74.55  \\
    CaLa~\cite{jiang2024cala}~\textcolor{gray}{(SIGIR'24)} &49.11 &	81.21 &	89.59 &	98.00 &	76.27 	&91.04 &	96.46 &	78.74  \\

    CoVR-2~\cite{ventura2024covr}~\textcolor{gray}{(TPAMI'24)} &50.43 &81.08& 88.89& 98.05& 76.75& 90.34& 95.78 & 79.28\\
    SPRC~\cite{bai2023sentence}~\textcolor{gray}{(ICLR'24)}& 51.96 &	82.12 &	89.74 &	97.69 &	\underline{80.65} &	92.31 &	96.60 	& 81.39  \\
    ENCODER~\cite{li2025encoder}~\textcolor{gray}{(AAAI'25)}& 46.10& 77.98& 87.16 &97.64& 76.92& 90.41& 95.95 &77.45\\
    CIR-LVLM~\cite{sun2025leveraging}~\textcolor{gray}{(AAAI'25)}& 53.64& 83.76 &  90.60  &97.93& 79.12& 92.33& 96.67 &81.44\\
    CCIN~\cite{tian2025ccin}~\textcolor{gray}{(CVPR'25)}& 53.41& 84.05 &  91.17  &98.00& - & - & - & - \\
    TME~\cite{li2025learning}~\textcolor{gray}{(CVPR'25)}& 53.42& 82.99 &90.24 &98.15 &\textbf{81.04} &92.58 &96.94 &82.01 \\
    QuRe~\cite{kwak2025qure}~\textcolor{gray}{(ICML'25)}& 52.22& 82.53 &  90.31  &98.17& 78.51 & 91.28 & 96.48 & 80.52 \\
    \rowcolor{gray!10}
    \multicolumn{1}{l}{\textbf{CIR-CoT (Fast)}}  & \underline{54.31} &	\underline{85.04} &	\underline{92.15} &	\underline{98.45} &	79.35 &	\underline{92.46} 	&\underline{97.30} & \underline{82.19}  \\
    \rowcolor[rgb]{ .851,  .851,  .851}
    \multicolumn{1}{l}{\textbf{CIR-CoT (Full)}}  & \textbf{55.06} &	\textbf{85.47} &	\textbf{92.60} &	\textbf{98.53} &	79.52 &	\textbf{92.92} 	&\textbf{97.37} & \textbf{82.49}  \\
\bottomrule
    \end{tabular}
\vspace{-1.2em}
\label{tab:cirr}
\end{table*}

\subsection{Results on Fashion-IQ}  
Table~\ref{tab:fiq} reports the results on the Fashion-IQ benchmark. CIR-CoT achieves the best overall performance, with average scores of 56.29 R@10 and 76.42 R@50, surpassing all prior methods. In particular, our model outperforms strong recent approaches such as CIR-LVLM, CCIN, and TME, showing consistent gains across all three categories. Notably, CIR-CoT delivers the highest R@10 in Dresses and Tops\&Tees, highlighting its effectiveness in handling fine-grained compositional queries in the fashion domain. Although CIR-CoT consistently achieves the best overall results on Fashion-IQ, the margin over recent strong baselines such as CIR-LVLM and TME is smaller compared to the clear advantage observed on CIRR. This is mainly because Fashion-IQ contains domain-specific fashion items with limited visual diversity and relatively simple textual modifications. As a result, the benefit of our CoT-enhanced reasoning and semantic compression is less pronounced compared to the more complex and diverse scenarios in CIRR, where fine-grained reasoning plays a larger role.

\subsection{Results on CIRCO} 
Table~\ref{tab:circo} reports the zero-shot evaluation results on the CIRCO dataset. 
In this setting, supervised methods are first trained on the CIRR dataset and then directly tested on CIRCO test set, which serves as a benchmark to assess cross-domain adaptability. 
In contrast, unsupervised approaches do not require additional training and can be directly evaluated on CIRCO.

Among the supervised group, CIR-CoT achieves a significant performance gain. 
For instance, it reaches an mAP@5 of 33.54, outperforming the previous best supervised method SPRC~\cite{bai2023sentence} (22.86) by +10.68 points. 
This advantage is consistent across other cutoffs, with CIR-CoT obtaining 37.29 mAP@50 compared to 26.55 for SPRC. 
Even when compared with the strongest unsupervised method OSrCIR~\cite{Tang2024OR}, which achieves 36.59 mAP@50, CIR-CoT still surpasses it by +0.70 while showing a much larger improvement at lower ranks.  
These results demonstrate that CIR-CoT, by incorporating structured chain-of-thought reasoning, not only excels in in-domain retrieval but also generalizes across domains, achieving state-of-the-art zero-shot performance on CIRCO.
\begin{table*}[]
  \centering
  \setlength{\tabcolsep}{11pt}
  \vspace{-10pt}
\caption{Performance comparison on Fashion-IQ validation set in terms of R@$k$ (\%).}
  \vspace{-10pt}
    \small
    \begin{tabular}{lcccccccc}
    \toprule
    \multicolumn{1}{c}{\multirow{2}{*}{Method}} & \multicolumn{2}{c}{Dresses} & \multicolumn{2}{c}{Shirts} & \multicolumn{2}{c}{Tops\&Tees} & \multicolumn{2}{c}{Avg} \\
    \cmidrule(lr){2-3}\cmidrule(lr){4-5}\cmidrule(lr){6-7}\cmidrule(lr){8-9}
    & R@10 & R@50 & R@10 & R@50 & R@10 & R@50 & R@10 & R@50 \\
    \midrule
    MGUR~\cite{Chen2022ComposedIR}~\textcolor{gray}{(ICLR'24)}& 32.61 & 61.34 & 33.23 & 62.55 & 41.40 & 72.51 & 35.75 & 65.47 \\
    FashionSAP~\cite{Han2023FashionSAPSA}~\textcolor{gray}{(CVPR'23)} & 33.71 & 60.43 & 41.91 & 70.93 & 33.17 & 61.33 & 36.26 & 64.23 \\
    FAME-ViL~\cite{Han2023FAMEViLMV}~\textcolor{gray}{(CVPR'23)} & 42.19 & 67.38 & 47.64 & 68.79 & 50.69 & 73.07 & 46.84 & 69.75 \\
    SyncMask~\cite{Song2024SyncMaskSA}~\textcolor{gray}{(CVPR'24)}& 33.76 & 61.23 & 35.82 & 62.12 & 44.82 & 72.06 & 38.13 & 65.14 \\
    SADN~\cite{wang2024semantic}~\textcolor{gray}{(ACM MM'24)} &40.01 & 65.10 & 43.67 & 66.05 & 48.04 & 70.93 & 43.91 & 67.36 \\
    CaLa~\cite{jiang2024cala}~\textcolor{gray}{(SIGIR'24)} &42.38& 66.08& 46.76& 68.16 &50.93& 73.42 &46.69 & 69.22 \\
    CoVR-2~\cite{ventura2024covr}~\textcolor{gray}{(TPAMI'24)} &46.53& 69.60 &51.23 &70.64& 52.14 &73.27& 49.96 &71.17\\
    SPRC~\cite{bai2023sentence}~\textcolor{gray}{(ICLR'24)}& 49.18 & 72.43 & 55.64 & 73.89 & 59.35 & 78.58 & 54.72 & 74.97 \\
    FashionERN~\cite{chen2024fashionern}~\textcolor{gray}{(AAAI'24)} & 50.32 & 71.29 & 50.15 & 70.36 & 56.40 & 77.21 & 52.29 & 72.95 \\
    CIR-LVLM~\cite{sun2025leveraging}~\textcolor{gray}{(AAAI'25)}&\underline{50.42} &\underline{73.60} &\textbf{58.59} &\textbf{75.86} &\underline{59.61} &\textbf{78.99} &\underline{56.21} &\underline{76.14}\\
    CCIN~\cite{tian2025ccin}~\textcolor{gray}{(CVPR'25)}&49.38 &72.58 &55.93 &74.14 &57.93 &77.56 &54.41 &74.76 \\
    TME~\cite{li2025learning}~\textcolor{gray}{(CVPR'25)}& 49.73 & 71.69 &56.43 &74.44 &59.31 &78.94 &55.15 &75.02 \\
    QuRe~\cite{kwak2025qure}~\textcolor{gray}{(ICML'25)} &46.80 &69.81 &53.53 &72.87 &57.47 &77.77 &52.60 &73.48\\
    \rowcolor[rgb]{ .851, .851, .851}
    \multicolumn{1}{l}{\textbf{CIR-CoT(Ours)}} & \textbf{50.82} & \textbf{74.57} & \underline{57.26} & \underline{75.76} & \textbf{60.79} & \underline{78.94} & \textbf{56.29} & \textbf{76.42} \\
    \bottomrule
    \end{tabular}
  \label{tab:fiq}
  \vspace{-1em} 
\end{table*}

\begin{table}[]
\caption{Zero-shot CIR performance on the CIRCO~\cite{Baldrati2023ZeroShotCI} test set.}
\label{tab:circo}
\centering
\vspace{-10pt}
\resizebox{1.0\linewidth}{!}{
\begin{tabular}{cc cccc}
\toprule
\multirow{2}{*}{\textbf{Method}} & \multirow{2}{*}{\textbf{Supervised}} & \multicolumn{4}{c}{\textbf{mAP@k}} \\
\cmidrule(lr){3-6}
 & & \textbf{k=5} & \textbf{k=10} & \textbf{k=25} & \textbf{k=50} \\
\midrule
CompoDiff~\cite{Gu2023CompoDiffVC}~\textcolor{gray}{(TMLR'24)} & \textcolor{red}{\XSolidBrush} & 15.30 & 17.70 & 19.50 & 21.00 \\
LinCIR~\cite{Gu2023LanguageonlyET}~\textcolor{gray}{(CVPR'24)} & \textcolor{red}{\XSolidBrush} & 19.71 & 21.01 & 23.13 & 24.18 \\
CIReVL~\cite{Karthik2023VisionbyLanguageFT}~\textcolor{gray}{(ICLR'24)} & \textcolor{red}{\XSolidBrush} &27.12 &28.01 &30.35 &31.39 \\
PrediCIR~\cite{Tang2025MissingTI}~\textcolor{gray}{(CVPR'25)} & \textcolor{red}{\XSolidBrush} & 23.70 & 24.60 & 25.40 & 26.00 \\
OSrCIR~\cite{Tang2024OR}~\textcolor{gray}{(CVPR'25)} & \textcolor{red}{\XSolidBrush} &30.47 &31.14 &35.03 &36.59 \\
\midrule
Q-Former~\cite{Li2023BLIP2BL} & \textcolor{teal}{\Checkmark} & 17.50 & 19.20 & 21.00 & 22.30 \\
SPRC~\cite{bai2023sentence}~\textcolor{gray}{(ICLR'24)} & \textcolor{teal}{\Checkmark} &22.86 &23.63 &25.56 &26.55 \\
\rowcolor[rgb]{ .851, .851, .851}
\textbf{CIR-CoT(Ours)} & \textcolor{teal}{\Checkmark} &\textbf{33.54} &\textbf{34.11} &\textbf{36.29} &\textbf{37.29} \\
\bottomrule
\end{tabular}
}
\vspace{-1em}
\end{table}

\subsection{Ablation Study}


\noindent\textbf{Study on the core components.}  
We analyze three components of CIR-CoT: stage-1 pretraining, CoT-augmented data, and training the vision projector (V.P.). As shown in Table~\ref{tab:ab_core}, directly fine-tuning from Qwen2.5-VL gives 52.68 R@1. Adding CoT data or stage-1 pretraining individually brings modest gains (+0.24 and +0.34 R@1). With both, even a frozen projector achieves 54.21 R@1, already outperforming previous settings. Jointly training the projector (\emph{CIR-CoT (Full)}) further improves performance, yielding +2.38, +2.48, and +2.39 gains at R@1, R@5, and R@10 over the baseline, confirming the complementary benefits of all components.

\noindent\textbf{Study on the loss weighting coefficients.}  
Table~\ref{tab:ab_lambda} investigates the influence of the weighting coefficient $\lambda_{txt}$ for the text generation loss while keeping $\lambda_{\text{Info}}$ fixed at $1.0$. We observe that setting $\lambda_{txt}=1.0$ yields the best overall performance, reaching 55.06 R@1 and 85.47 R@5. When $\lambda_{txt}$ is too small (e.g., $0.5$ or $0.7$), the model underperforms due to insufficient supervision from the text generation objective. Conversely, increasing $\lambda_{txt}$ beyond $1.0$ (e.g., $1.5$ or $2.0$) also degrades performance, because the model overemphasizes text generation at the expense of retrieval alignment. These results highlight that a balanced weighting between the text generation loss and the InfoNCE loss is crucial for optimizing retrieval effectiveness.

\begin{table}[]
\centering
\caption{Ablation Study of Components on the CIRR Dataset. The V.P. stands for Vision Projector.}
\label{tab:ab_core}
\vspace{-1em}
\resizebox{1.0\linewidth}{!}{
\begin{tabular}{cc c cccc}
\toprule
\multirow{2}{*}{\textbf{Method}} & \multirow{2}{*}{Stage 1} & \multirow{2}{*}{CoT data} & \multicolumn{4}{c}{R@K} \\
\cmidrule(lr){4-7}
& & & \textbf{K=1} & \textbf{K=5} & \textbf{K=10} & \textbf{K=50} \\
\midrule
Baseline & \textcolor{red}{\XSolidBrush} & \textcolor{red}{\XSolidBrush} & 52.68 & 82.99 & 90.21 & 97.45 \\
Base + CoT & \textcolor{red}{\XSolidBrush} & \textcolor{teal}{\Checkmark} & 52.92 & 83.13 & 91.48 & 98.43 \\
Base + Stage 1 & \textcolor{teal}{\Checkmark} & \textcolor{red}{\XSolidBrush} & 53.02 & 84.06 & 91.99 & 98.42 \\
\midrule
CIR-CoT (frozen V.P.) & \textcolor{teal}{\Checkmark} & \textcolor{teal}{\Checkmark} &54.21 &85.06 &92.14 &98.44\\
\rowcolor{gray!18}
\textbf{CIR-CoT (Full)} & \textcolor{teal}{\Checkmark} & \textcolor{teal}{\Checkmark} & \textbf{55.06} & \textbf{85.47} & \textbf{92.60} & \textbf{98.53} \\
\bottomrule
\end{tabular}}
\vspace{-1.0em}
\end{table}

\begin{table}[]
\centering
\setlength{\tabcolsep}{10pt}
\caption{Ablation Study on Loss Weighting Coefficients.}
\label{tab:ab_lambda}
\vspace{-1em}
\small
\begin{tabular}{cc cccc}
\toprule
\multirow{2}{*}{\textbf{$\lambda_{txt}$}} & \multirow{2}{*}{$\lambda_{\text{Info}}$} & \multicolumn{4}{c}{\textbf{R@K}} \\
\cmidrule(lr){3-6}
& & \textbf{K=1} & \textbf{K=5} & \textbf{K=10} & \textbf{K=50} \\
\midrule
0.5 & 1.0 & 53.14 & 83.78 & 91.56 & 98.28 \\
0.7 & 1.0 & 54.23 & 84.57 & 92.19 & 98.36 \\
\rowcolor{gray!18}
\textbf{1.0} & \textbf{1.0} & \textbf{55.06} & \textbf{85.47} & \textbf{92.60} & \textbf{98.53} \\
1.5 & 1.0 & 53.45 & 83.98 & 91.18 & 98.43 \\
2.0 & 1.0 & 53.02 & 84.05 & 91.25 & 98.24 \\
\bottomrule
\end{tabular}
\vspace{-1em}
\end{table}

\subsection{Qualitative Results}  
Fig.~\ref{fig:visual} presents qualitative comparisons on the CIRR dataset.  
Unlike traditional retrieval models that directly match queries with images, our CIR-CoT explicitly performs step-by-step reasoning to interpret the user’s modification instruction and generate an intermediate description of the target image.  
This reasoning process enables the model to focus on the required changes, such as inserting a tent interior with a black pole or modifying the pose and accessories of humans in underwater scenes, while preserving irrelevant details.  
As a result, CIR-CoT produces more faithful and semantically aligned retrieval outcomes, which baseline methods like SPRC fail to capture. 
More detailed qualitative analyses and additional examples are provided in the supplementary material.

\section{Conclusion}
In this work, we presented \textbf{CIR-CoT}, a framework that leverages chain-of-thought reasoning to enhance image retrieval from natural language queries. CIR-CoT performs explicit reasoning and generates interpretable intermediate descriptions before final retrieval. Experiments on various CIR datasets demonstrate that our approach achieves competitive performance and strong cross-domain adaptability. Qualitative results further show that CIR-CoT can infer user intent and articulate target image descriptions, surpassing traditional CIR methods. This work paves the way for integrating reasoning into multimodal retrieval and provides a foundation for developing more interpretable CIR systems.

{
    \small
    \bibliographystyle{ieeenat_fullname}
    \bibliography{main}
}

\end{document}